\documentclass[conference,letterpaper]{IEEEtran}
\IEEEoverridecommandlockouts
\usepackage{amsmath}
\usepackage{mathrsfs}
\usepackage{url}
\usepackage{csquotes}

\usepackage[dvips]{graphicx}  

\usepackage{multirow}
\usepackage[left=0.71in,top=0.94in,right=0.71in,bottom=1.18in]{geometry}
\usepackage[colorinlistoftodos]{todonotes}
\usepackage{xcolor}
\usepackage{soul}

\sethlcolor{green}

\begin{document}
\title{\huge Real-Time Subpixel Fast Bilateral Stereo}

\author{Rui Fan$^{1,2}$, Yanan Liu$^{3}$, Mohammud Junaid Bocus$^{2}$, Lujia Wang$^{4}$, Ming Liu$^{1}$\\
	{$^{1}$Robotics and Multi-Perception Laborotary, }\\
	{Department of Electronic and Computer Engineering,}\\
	{The Hong Kong University of Science and Technology, Hong Kong SAR, China.}
	\\
	{$^{2}$Visual Information Institute, University of Bristol, Bristol, BS8 1UB, United Kingdom.}
	\\
{$^{3}$Bristol Robotics Laboratory, University of Bristol, Bristol, BS16 1QY, United Kingdom.}
\\
{$^{4}$Shenzhen Institutes of Advanced Technology, Chinese Academy of Sciences, Shenzhen, China.}
\\
}



\maketitle
\begin{abstract}
Stereo vision technique has been widely used in robotic systems to acquire 3-D information. 
In recent years, many researchers have applied bilateral filtering in stereo vision to adaptively aggregate the matching costs. This has greatly improved the accuracy of the estimated disparity maps. However, the process of filtering the whole cost volume is very time consuming and therefore the researchers have to resort to some powerful hardware for the real-time purpose.  This paper presents the implementation of fast bilateral stereo on a state-of-the-art GPU. By highly exploiting the parallel computing architecture of the GPU, the fast bilateral stereo performs in real time when processing the Middlebury stereo datasets.

\end{abstract}

\section{Introduction}
\label{sec.introduction}

We live in a 3-D world but our eyes can only perceive objects in 2-D. The miracle of our depth perception is due to our brain's ability to analyse the difference between the two 2-D images which are projected on the retinas of our eyes \cite{Qian1997}. As for the digital images captured by cameras, they are 2-D in nature. By comparing the difference between the images captured using a pair of synchronised cameras, the 3-D information of the surrounding environment can be obtained \cite{Hartley2003}. This process is commonly referred to as \textit{stereo vision}, and it is very similar to the human binocular vision \cite{Emanuele1998}. For a well-calibrated stereo rig, the horizontal distance between each pair of corresponding points  $\boldsymbol{p_l}=[u_l,v_l]^\top$ in the left image $\pi_l$ and $\boldsymbol{p_r}=[u_r,v_r]^\top$ in the right image $\pi_r$ is  defined as \textit{disparity} \cite{Fan2018}.

The two important factors in stereo vision are speed and accuracy \cite{Fan2017}. A lot of research has been carried out over the past decade to improve both the precision of the disparity maps and the execution speed of the algorithms \cite{Tippetts2012, Tippetts2016}. {However,  the stereo vision algorithms designed to achieve better disparity accuracy usually have higher computational complexity and lower processing efficiency \cite{Yang2009}}. Hence, speed and accuracy are two desirable but conflicting properties, and it is very challenging to achieve both of them simultaneously \cite{Tippetts2016}. Therefore, the main motivation of developing a stereo vision algorithm is to improve the trade-off between accuracy and speed \cite{Tippetts2016}. In most circumstances, a desirable trade-off entirely depends on the target application. For example, {a} real-time performance is required {for} the stereo vision systems applied in mobile robotics because the other systems, e.g., lane detection \cite{Ozgunalp2017}, visual odometry \cite{Evans2018}, visual Simultaneous Localisation and Mapping (SLAM) \cite{Yun2017, Jiao2017} and visual tracking \cite{Pomerleau2011, Liao2017}, can be easily implemented in real time if the 3-D information is available \cite{Fan2016}. 

\begin{figure*}[!t]
	\begin{center}
		\centering
		\includegraphics[width=1\textwidth]{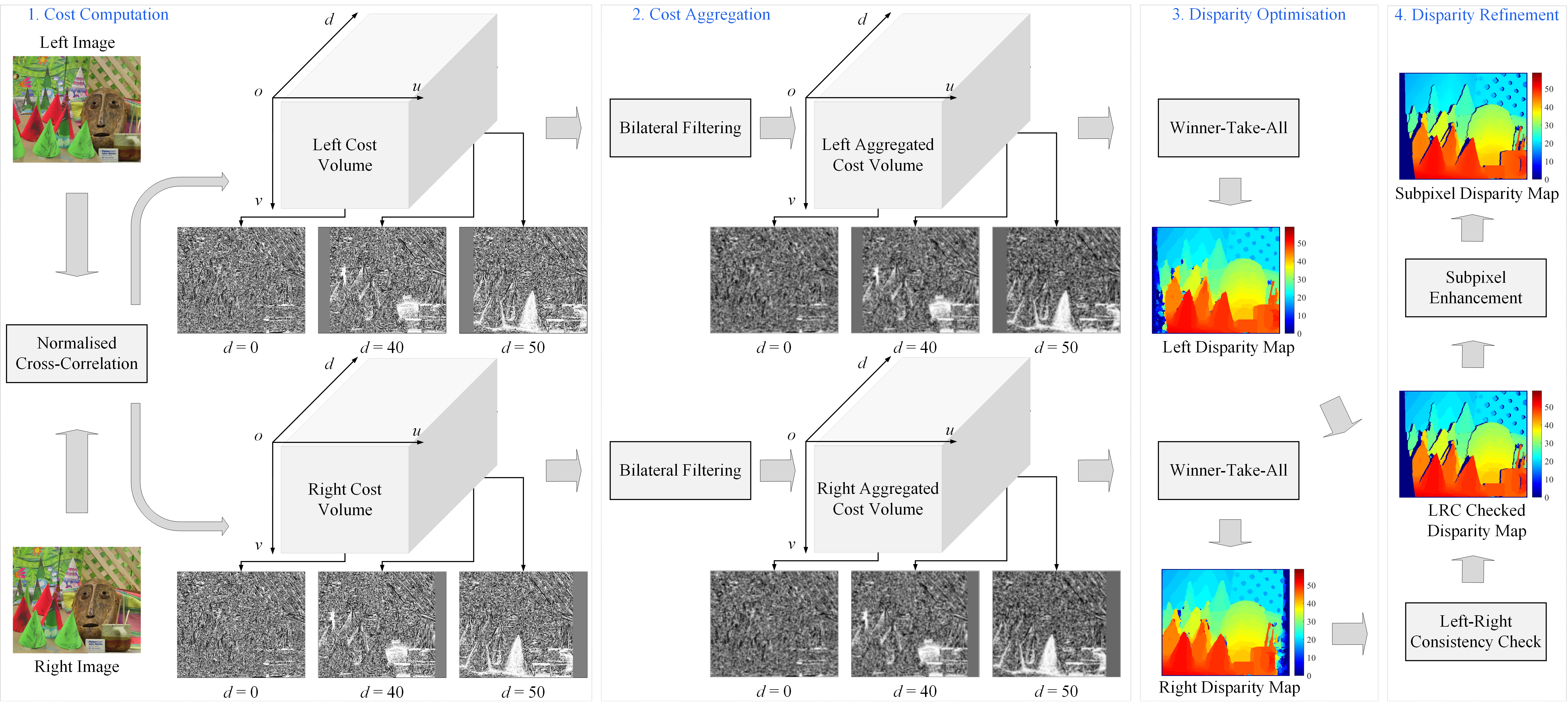}
		\caption{Overview of the proposed disparity estimation system.}
		\label{fig.block_diagram}
	\end{center}
\end{figure*}

The state-of-the-art algorithms for disparity estimation can mainly be classified as local and global \cite{Fan2018}.  The local algorithms \cite{Sara2002, Sara2006, Cech2007, Fan2017} simply match a series of blocks from the left and right images and select the correspondence with the lowest cost or the highest correlation. This optimisation is also known as Winner-Take-All (WTA) \cite{Fan2017}. Unlike the local algorithms, the global algorithms process the stereo matching using some sophisticated techniques, e.g., Belief Propagation (BP) \cite{Ihler2005} and Graph Cut (GC) \cite{Boykov2001}. These techniques are usually developed based on the Markov Random Fields (MRF) \cite{Tappen2003}, where the process of finding the best disparities translates to a probability maximisation problem \cite{Fan2018}. However, finding the optimum values for the smoothness parameters is a difficult task due to the occlusion problem. Over-penalising the smoothness term can reduce the ambiguities around the discontinuities but on the other hand can cause errors for continuous areas \cite{Fan2018}. In \cite{Mozerov2015}, Mozerov and Weijer proved that the initial energy optimisation problem in a fully connected MRF model can be formulated as an adaptive cost aggregation using bilateral filtering. Since then, a lot of endeavours have been made in local algorithms to improve the accuracy of the disparity maps by performing bilateral filtering on the cost volume before estimating the disparities. These algorithms are also known as fast bilateral stereo (FBS). However, filtering the whole cost volume is always computationally intensive  and therefore the FBS has to be implemented on some powerful hardware for the purpose of real-time  \cite{Fan2018}. 

This paper presents the implementation of the FBS on an NVIDIA GTX 1080 GPU. Firstly, we provide the readers with some mathematical preliminaries about the FBS. The cost computation is optimised based on our previously published algorithm in \cite{Fan2017}. Furthermore, each matching cost is saved in two 3-D cost volumes, hence reducing the  computations by $50\%$ when estimating both the left and right disparity maps. Then, we provide some details on the practical implementation, such as the use of different types of device memory. The source code of our implementation is publicly available at: \url{https://bitbucket.org/rangerfan/bilateral_stereo_gpu_middlebury.git}. 

The remainder of this paper is organised as follows: Section \ref{sec.algorithm_description} describes the proposed disparity estimation algorithm.  Implementation details are provided in Section \ref{sec.implementation}. In Section \ref{sec.exp_results}, we present the experimental results and evaluate the performance of the proposed stereo vision system. Section \ref{sec.conclusion_fw} summaries the paper and provides some recommendations for future work.  

\section{Algorithm Description}
\label{sec.algorithm_description}

In general, a disparity estimation algorithm usually consists of four steps: cost computation, cost aggregation, disparity optimisation and disparity refinement \cite{Mozerov2015}. {However, the sequential use of these steps entirely depends on the chosen algorithm \cite{Scharstein2002}. In the following subsections, we discuss each step in the proposed disparity estimation algorithm.}


\subsection{Cost Computation}
\label{sec.cost_computation}

In stereo matching, the most commonly used approaches for cost computation are the Absolute Difference (AD) and the Squared Difference (SD) \cite{Scharstein2011}. However, these two approaches are very sensitive to the intensity difference between the left and right images, which may further lead to some incorrect matches in the process of disparity estimation. In this paper, we use the Normalised Cross-Correlation (NCC) as the cost function to measure the similarity between each pair of blocks selected from the left and right images. Although the NCC is more computationally intensive than the AD and  the SD, it can provide more accurate results when the intensity difference is involved \cite{Fan2018}. The cost function of the NCC is as follows \cite{Fan2017}:
\begin{equation}
c(u,v,d)=\frac{\sum\limits_{x=u-\varrho}^{x=u+\varrho}\sum\limits_{y=v-\varrho}^{y=v+\varrho} i_{l}(x,y) i_{r}(x-d,y)-n\mu_{l} \mu_{r}}{n\sigma_l \sigma_r}
\label{eq.ncc}
\end{equation}
where
\begin{equation}
\sigma_{l}=\sqrt{\sum\limits_{x=u-\varrho}^{x=u+\varrho}\sum\limits_{y=v-\varrho}^{y=v+\varrho}{i_{l}}^2(x,y)/n-{\mu_{l}}^2}
\label{eq.sigma_l_1}
\end{equation}
\begin{equation}
\sigma_{r}=\sqrt{\sum\limits_{x=u-\varrho}^{x=u+\varrho}\sum\limits_{y=v-\varrho}^{y=v+\varrho}{i_{r}}^2(x-d,y)/n-{\mu_{r}}^2}
\label{eq.sigma_r_1}
\end{equation}


$c(u,v,d)\in[-1,1]$ is the correlation cost. $i_l(x,y)$ denotes the intensity of a pixel at $(x,y)$ in the left image and $i_r(x-d,y)$ represents the intensity of a pixel at $(x-d,y)$ in the right image. The centre of the reference square block is $(u,v)$. Its width is $2\varrho+1$ and the number of pixels within each block is $n=(2\varrho+1)^2$. $\varrho$ is set to $1$ in this paper. $\mu_l$ and $\mu_r$ represent the means of the pixel intensities within the left and right blocks, respectively. $\sigma_l$ and $\sigma_r$ denote the standard deviations of the left and right blocks, respectively.


In practical implementation, the values of $\mu_l$, $\mu_r$, $\sigma_l$ and $\sigma_r$ are pre-calculated and stored in a static program storage for direct indexing \cite{Fan2018}. Therefore, only the dot product $\sum i_l i_r$ in Eq. (\ref{eq.ncc}) {needs to be calculated when computing the correlation cost between each pair of left and right blocks}. This greatly reduces unnecessary computations. More details on the NCC implementation are provided in our previously published work \cite{Fan2017}.

The calculated correlation costs $c$ are simultaneously stored in the left and right 3-D cost volumes, as shown in Fig. \ref{fig.block_diagram}. It is to be noted that the value of $c$ at the position of $(u,v,d)$ in the left cost volume is the same as that at the position of $(u-d,v,d)$ in the right cost volume. In the next step, each correlation cost in the two 3-D cost volumes is updated by aggregating the costs from its neighbourhood system.

\subsection{Cost Aggregation}
\label{sec.cost_aggregation}

In global algorithms, finding the best disparities is equivalent to {maximising} the joint probability in Eq. (\ref{eq.mrf_eq1}) \cite{Fan2018}.

%

\begin{equation}
P(\boldsymbol{p}, q)=\prod_{\boldsymbol{p_{ij}}\in\mathscr{P}} \Phi(\boldsymbol{p_{ij}}, q_{\boldsymbol{p_{ij}}})\prod_{\boldsymbol{n_{p_{ij}}}\in\mathscr{N}_{\boldsymbol{p_{ij}}}} \Psi (\boldsymbol{p_{ij}}, \boldsymbol{n_{p_{ij}}})
\label{eq.mrf_eq1}
\end{equation}
where $\boldsymbol{p_{ij}}$ represents a vertex at the site of $(i,j)$ in the disparity map $\mathscr{P}$ and $q_{\boldsymbol{p_{ij}}}$ denotes the intensity differences which correspond to different disparities $d$. $\mathscr{N}_{\boldsymbol{p_{ij}}}=\{n_{1\boldsymbol{p_{ij}}},n_{2\boldsymbol{p_{ij}}},n_{3\boldsymbol{p_{ij}}},\cdots,n_{k\boldsymbol{p_{ij}}}|n_{\boldsymbol{p_{ij}}}\in\mathscr{P}\}$ represents the neighbourhood system {of} $\boldsymbol{p_{ij}}$. In this paper, the value of $k$ is set to $8$ and $\mathscr{N}$ is an 8-connected neighbourhood. $\Phi(\cdot)$ expresses the compatibility between  each possible disparity $d$ and the corresponding block similarity. $\Psi(\cdot)$ expresses the compatibility between $\boldsymbol{p_{ij}}$ and its neighbourhood system $\mathscr{N}_{\boldsymbol{p}_{ij}}$. However, Eq. (\ref{eq.mrf_eq1}) is commonly formulated  as an energy function, as follows \cite{Fan2018}:

\begin{equation}
\begin{split}
E(\boldsymbol{p})&=\sum_{\boldsymbol{p_{ij}}\in\mathscr{P}} D(\boldsymbol{p_{ij}}, q_{\boldsymbol{p_{ij}}})+
\sum_{\boldsymbol{n_{p_{ij}}}\in\mathscr{N}_{\boldsymbol{p_{ij}}}} V (\boldsymbol{p_{ij}}, \boldsymbol{n_{p_{ij}}})\\
\end{split}
\label{eq.mrf_eq2}   
\end{equation}
where $D(\cdot)$ corresponds to the correlation cost $c$ in this paper and $V(\cdot)$ determines the aggregation strategy. In the MRF model, the method to formulate an adaptive $V(\cdot)$ is important because the intensity in a discontinuous area usually greatly differs from {those} of its neighbours \cite{Li2012}. Therefore, some authors formulated $V(\cdot)$ as a piece-wise model to discriminate the discontinuous areas \cite{Kolmogorov2001}. However, the process of minimising the energy function in Eq. (\ref{eq.mrf_eq2}) results in a high computational complexity, making real-time performance challenging.
Since Tomasi et al. introduced the bilateral filter in \cite{Tomasi1998}, many authors have investigated its application to aggregate the matching costs \cite{Yang2009, Hosni2013, Zhang2013a}. Mozerov and Weijer also proved that the bilateral filtering is a feasible solution for the energy minimisation problem in the MRF model \cite{Mozerov2015}. These methods are also known as fast bilateral stereo, where both  intensity difference and spatial distance provide a Gaussian weighting function to adaptively constrain the cost aggregation from the neighbours. A general representation of the cost aggregation in the FBS is as follows:

\begin{equation}
c_{agg}(u,v,d)=\frac{\sum_{x=u-\rho}^{x=u+\rho}  \sum_{y=v-\rho}^{y=v+\rho}  \omega_d(x,y)\omega_r(x,y)c(x,y,d)}{\sum_{x=u-\rho}^{x=u+\rho}  \sum_{y=v-\rho}^{y=v+\rho} \omega_d(x,y)\omega_r(x,y)}
\label{eq.fbs}
\end{equation}
where 
\begin{equation}
\omega_d(x,y)=\exp \bigg\{ -\frac{(x-u)^2+(y-v)^2}{{\gamma_d}^2} \bigg\}
\label{eq.omega_d}
\end{equation}
\begin{equation}
\omega_r(x,y)=\exp \bigg\{ -\frac{(i(x,y)-i(u,v))^2}{{\gamma_r}^2} \bigg\}
\label{eq.omega_r}
\end{equation}

$\omega_d$ and $\omega_r$ are based on the spatial distance and the colour similarity, respectively. $\gamma_d$ and $\gamma_r$ are two parameters used to control the values of $\omega_d$ and $\omega_r$, respectively. The costs $c$ within a square block are aggregated adaptively and an updated cost $c_{agg}$ can be obtained.

Although the FBS has shown a good performance in terms of matching accuracy, it usually takes a long time to process the whole cost volume. Therefore, the FBS has to been implemented on some powerful hardware in order to perform in real time.

\subsection{Disparity Optimisation}
\label{sec.disp_optimisation}

As discussed in Section \ref{sec.cost_aggregation}, the process of energy minimisation in global algorithms can be realised by performing bilateral filtering on the initial cost volumes. The best disparities can therefore be found by simply performing WTA on the left and right aggregated cost volumes. The left and right disparity maps, i.e., $\ell^{lf}$ and $\ell^{rt}$, are shown in Fig. \ref{fig.block_diagram}.

\subsection{Disparity Refinement}
\label{sec.disp_refinement}

This step usually involves several disparity map post-processing algorithms, such as weighted median filtering \cite{Yang2009}, left-right consistency (LRC) check \cite{Mei2011} and subpixel enhancement  \cite{Fan2018}. In this subsection, we only discuss the latter two algorithms.  

Firstly, according to the uniqueness constraint stated in \cite{Fan2017}, for an arbitrary point $\boldsymbol{p_l}=[u_l,v_l]^\top$ in the left image $\pi_l$, there exists at most one corresponding point $\boldsymbol{p_r}=[u_r,v_r]^\top$ in the right image $\pi_r$, namely:
\begin{equation}
\ell^{lf}(u,v)=\ell^{rt}(u-\ell^{lf}(u,v),v)
\end{equation}

Therefore, the LRC check is always performed to remove the incorrect matches in the occlusion areas and find an outline in the disparity map \cite{Fan2017}. The corresponding LRC check result is illustrated in Fig. \ref{fig.block_diagram}.

\begin{figure*}[!t]
	\begin{center}
		\centering
		\includegraphics[width=1\textwidth]{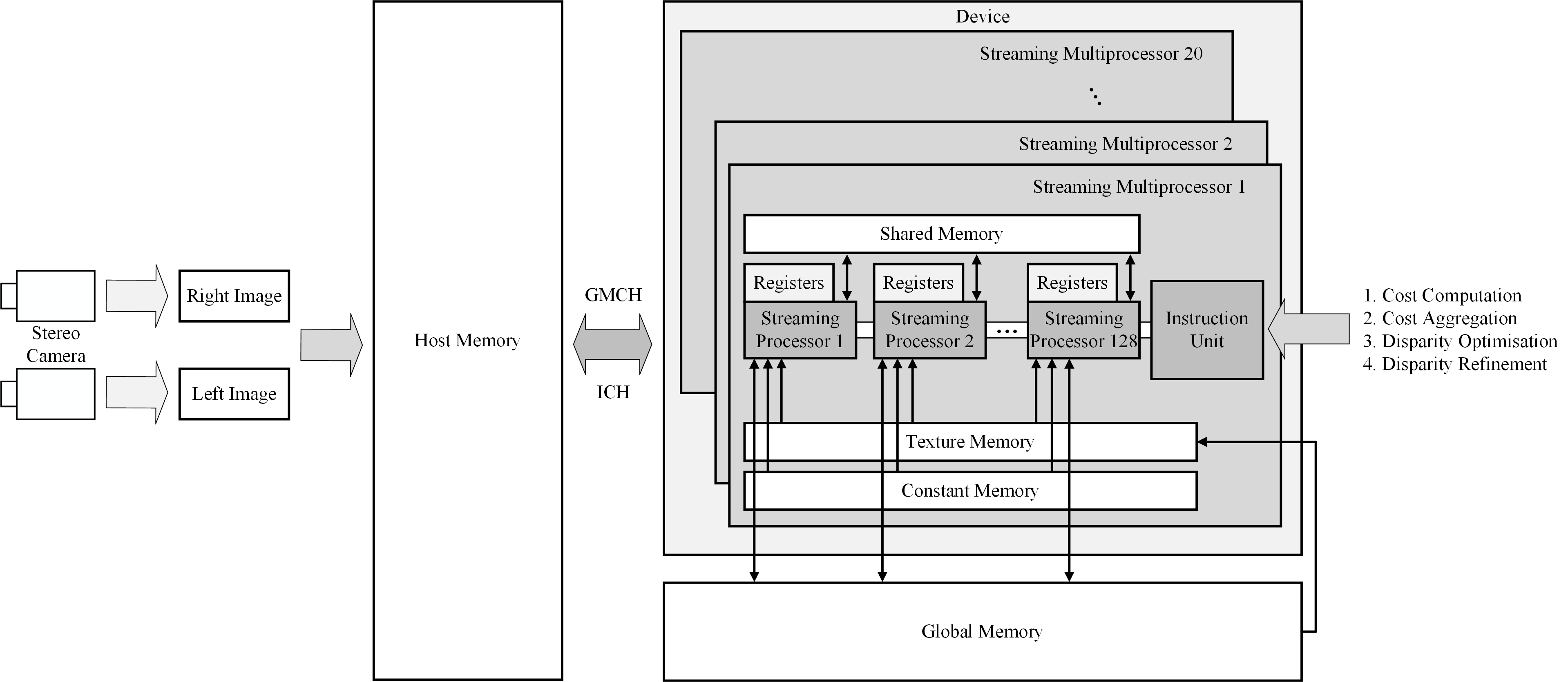}
		\caption{Overview of the practical implementation.}
		\label{fig.implementation}
	\end{center}
\end{figure*}

Furthermore, since the distance between a 3-D object and the camera focal point is inversely proportional to the disparity value, a disparity error larger than one pixel may result in a non-negligible difference in the 3-D reconstruction result \cite{Fan2018}. Therefore, a subpixel enhancement is always performed to increase the resolution of the disparity values. This can be achieved by fitting a parabola $f(u,v,d)$ to three correlation costs $c(u,v,d-1)$, $c(u,v,d)$ and $c(u,v,d+1)$ around the initial disparity $d$ and then selecting the centre line of $f(u,v,d)$ as the subpixel disparity $d^s$ \cite{Fan2018}: 

\begin{equation}
d^s=d+\frac{c(u,v,d-1)-c(u,v,d+1)}{2c(u,v,d-1)+2c(u,v,d+1)-4c(u,v,d)}
\label{eq.subpixel}
\end{equation}

The corresponding subpixel disparity map is shown in Fig. \ref{fig.block_diagram}. In the next section, we provide more details on the implementation. 

\section{Implementation}
\label{sec.implementation}

The proposed algorithm is implemented on an NVIDIA GTX 1080 GPU to achieve real-time performance. An overview of the practical implementation is shown in Fig. \ref{fig.implementation}. The GPU has 8 GB GDDR5X and 20 streaming multi-processors (SMs). Each SM consists of 128 streaming processors (SPs), and therefore the GPU has 2560 SPs in total. The Single Instruction Multiple Data (SIMD) architecture allows the SPs on the same SM to execute the same instruction but process different data at each clock cycle \cite{Fan2018a}. The host memory and the device memory are communicated with each other via Graphical/Memory Controller Hub (GMCH) and I/O Controller Hub (ICH), which are also known as the Intel northbridge and southbridge, respectively.

In our implementation, the left and right images are first sent to the global memory of the GPU from the host memory. Compared with the texture memory and constant memory which are read-only and cached on-chip, the global memory is off-chip and therefore less efficient in terms of memory requesting \cite{Fan2017}. Furthermore, a thread is more likely to fetch the data from the closest addresses that its nearby threads accessed, and thus the use of cache in global memory is impossible \cite{Fan2018a}. Hence, we use the texture memory to optimise the caching for the left and right images. We first create two 2-D texture objects on the texture memory. Then, the texture objects are bound directly to the  addresses of the left and right images in the global memory. The value of a pixel in the left or right image can thus be fetched from the texture objects instead of the global memory. This greatly minimises the memory requests from the global memory and further improves the speed of data fetching. Then, the instruction unit in each SM sends the same instructions to the SPs to compute and aggregate the correlation costs, and then optimise and refine the disparities.

In the first step, $\mu_l$, $\mu_r$, $\sigma_l$ and $\sigma_r$ are pre-calculated and their values are stored in global memory. When calculating the correlation cost using Eq. (\ref{eq.ncc}), the values of $\mu_l$, $\mu_r$, $\sigma_l$ and $\sigma_r$ are fetched from the global memory and the values of $i_l$ and $i_r$ are fetched from the texture memory. This significantly reduces the expensive computations of $\mu$ and $\sigma$ and the memory requests from the global memory. The value of each correlation cost $c$ is saved in two 3-D cost volumes in the global memory.

\begin{figure*}[!t]
	\begin{center}
		\centering
		\includegraphics[width=1\textwidth]{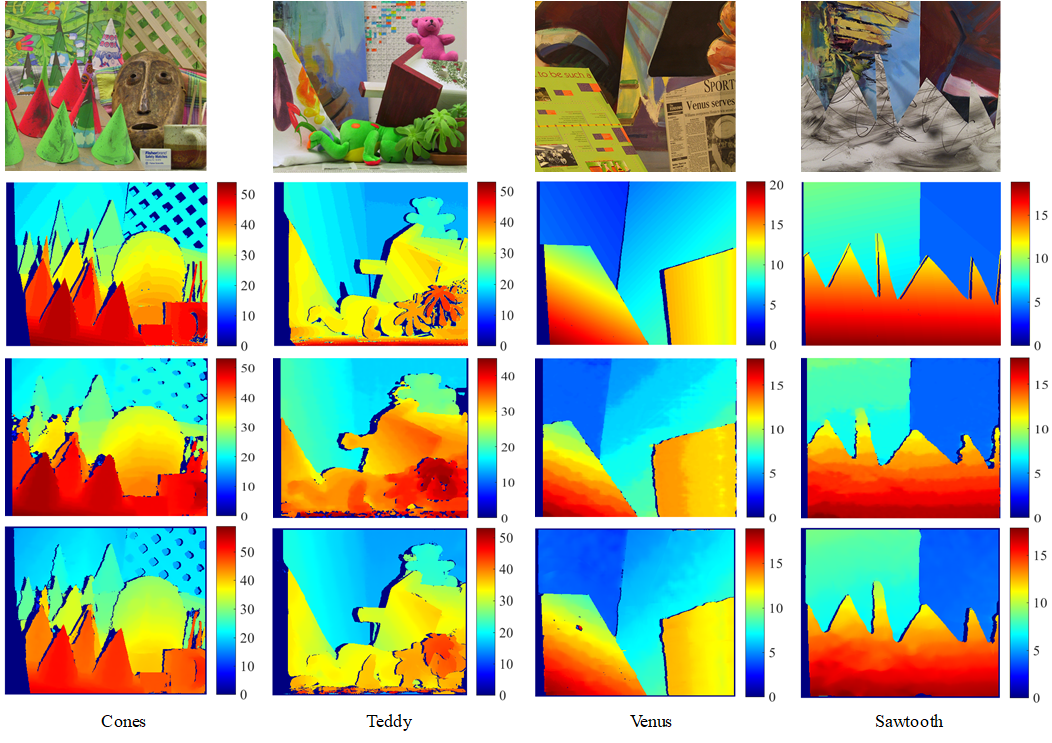}
		\caption{Experimental results. The first row illustrates the left images. The second row shows the ground truth disparity maps excluding the occlusion areas.  The third row shows the disparity maps estimated using semi-global matching.  The fourth row illustrates the experimental results.}
		\label{fig.experimental_results}
	\end{center}
\end{figure*}

As discussed in Section \ref{sec.cost_aggregation}, bilateral filtering is performed on the left and right cost volumes to aggregate the correlation costs adaptively. Due to the fact that constant memory is read-only and beneficial for the data that will not change over the course of a kernel execution, the values of $\omega_d$ and $\omega_r$ are pre-calculated and stored in the constant memory to reduce the unnecessary computations in Eq. (\ref{eq.omega_d}) and (\ref{eq.omega_r}). The left and right aggregated cost volumes are then stored in the global memory. 

In the third step, each SP searches for the highest correlation cost $c_{agg}$ between $(u,v,d_{min})$ and $(u,v,d_{max})$. The position in the $d$ axis which corresponds to the highest cost $c_{agg}$  is then selected as the desirable disparity for the position of $(u,v)$. The estimated left and right disparity maps are then stored in the global memory for the following processes. 

Finally, the instruction of disparity refinement is executed on each SM. In this paper, the left image is set as the reference, and therefore the LRC check is performed to remove the incorrect matches in the left disparity map. Then, each SP executes Eq. (\ref{eq.subpixel}) and stores the corresponding subpixel disparity values in the global memory. The subpixel disparity map is then sent to the host memory via GMCH and ICH for display. 

The performance of the proposed implementation is discussed in Section \ref{sec.exp_results}.

\section{Experimental Results}
\label{sec.exp_results}

As discussed in Section \ref{sec.introduction}, accuracy and speed are two main aspects of stereo vision and achieving desirable results  entirely depends on  a good trade-off between these two factors.  Hence in the following subsections, we evaluate both the precision of the proposed algorithm and the execution speed of the practical implementation.

\subsection{Accuracy Evaluation}
\label{sec.accuracy_eva}

In this subsection, we use Middlebury 2001 datasets \cite{Scharstein2002} and 2003 datasets \cite{Scharstein2003} to evaluate the accuracy of the proposed stereo matching algorithm. Some experimental results are illustrated in Fig. \ref{fig.experimental_results}.

To quantify the accuracy of the estimated disparity maps, Barron et al. proposed to compute the Percentage of Error Pixels $e_{PEP}$ as follows \cite{Barron1994}:

\begin{equation}
e_{PEP}={\frac{1}{N}\sum_{(u,v)}\delta\big(|\ell^{et}(u,v)-\ell^{gt}(u,v)|,\  \varepsilon_d\big)}\times 100\%
\label{eq.e_pep}
\end{equation}
where 
\begin{equation}
\delta(d, \varepsilon_d)=\bigg\{
\begin{aligned}
1 \ \ \ \ \ (d>\varepsilon_d)\\
0 \ \ \ \ \ (d\leq\varepsilon_d)
\end{aligned}
\end{equation}

\begin{figure}[!t]
	\begin{center}
		\centering
		\includegraphics[width=0.4\textwidth]{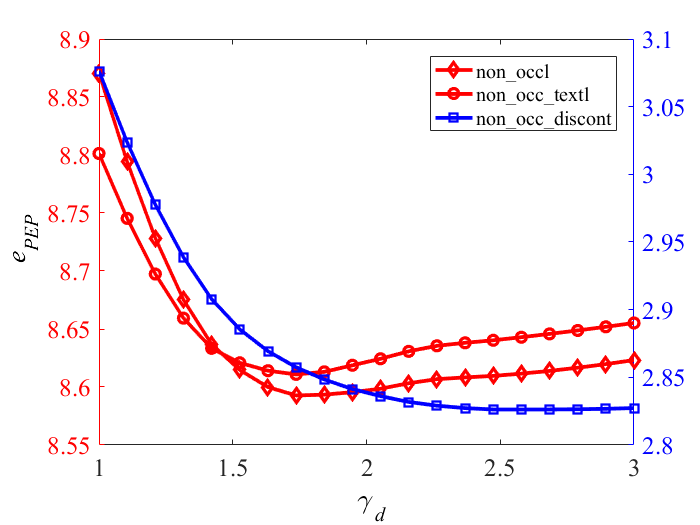}
		\caption{$e_{PEP}$ with respect to different values of $\gamma_d$.}
		\label{fig.eva1}
	\end{center}
\end{figure}
\begin{figure}[!t]
	\begin{center}
		\centering
		\includegraphics[width=0.4\textwidth]{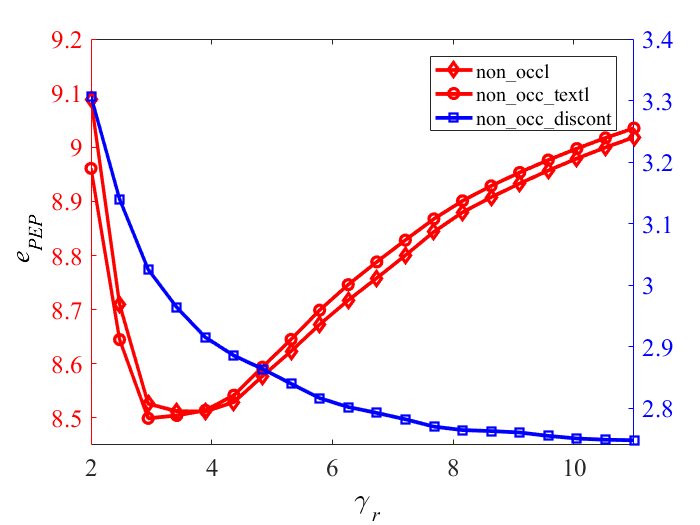}
		\caption{$e_{PEP}$ with respect to different values of $\gamma_r$.}
		\label{fig.eva2}
	\end{center}
\end{figure}
\begin{figure}[!t]
	\begin{center}
		\centering
		\includegraphics[width=0.4\textwidth]{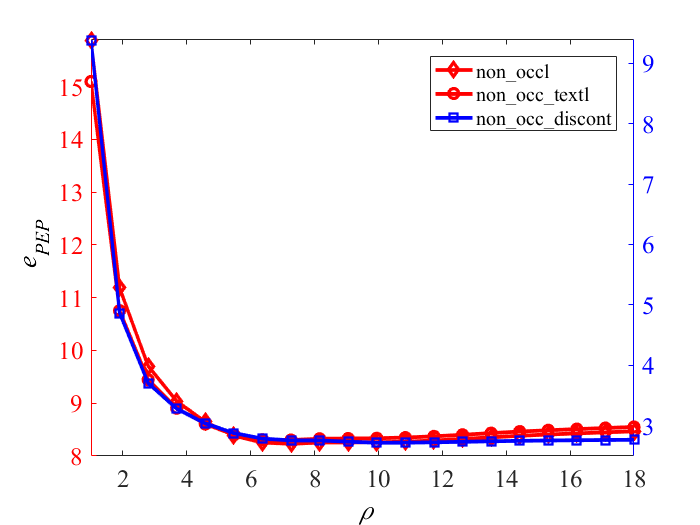}
		\caption{$e_{PEP}$ with respect to different values of $\rho$.}
		\label{fig.eva3}
	\end{center}
\end{figure}
\begin{figure}[!t]
	\begin{center}
		\centering
		\includegraphics[width=0.4\textwidth]{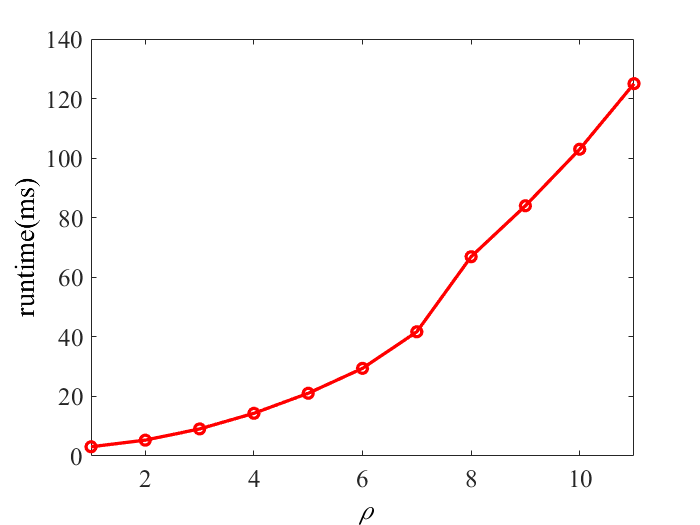}
		\caption{Runtime with respect to different values of $\rho$.}
		\label{fig.eva4}
	\end{center}
\end{figure}

The threshold $\varepsilon_d$ is used to determine whether an estimated disparity is correct or not and it is set to $2$ in this paper. $N$ stands for the total number of disparities used for evaluation. $\ell^{et}$ represents the estimated disparity map and $\ell^{gt}$ denotes the corresponding ground truth data. To provide an in-depth performance evaluation of the proposed algorithm, we compute the values of $e_{PEP}$ for different regions of interest: the disparity map excluding occlusions (non$\_$occ), the disparity map excluding occlusions and texture-less areas (non$\_$occ$\_$textl), and the disparity map excluding occlusions and discontinuities (non$\_$occ$\_$discont). The corresponding values of $e_{PEP}$ with respect to different input images and regions of interest are shown in Table \ref{table.e_pep}. Next, we use the \enquote{Cones} dataset to evaluate the performance of the algorithm with respect to different values of parameters in bilateral filtering, i.e., $\omega_d$, $\omega_r$ and $\rho$. The corresponding evaluation results are shown in Fig. \ref{fig.eva1}, \ref{fig.eva2} and \ref{fig.eva3}. 

\begin{table}[!b]
	\begin{center}
		\footnotesize
		\caption{$e_{PEP}$ with respect to different input images and regions of interest ($\varepsilon_d=2$).}
		\begin{tabular}{|c|c|c|c|c|c|c|}
			\hline
			Region of interest & Cones & Teddy & Venus & Sawtooth \\
			\hline
			non$\_$occl & 8.2264 & 10.9244  & 2.8573  & 7.3800 \\ 
			non$\_$occ$\_$textl  & 8.2928 &  10.8922 &  n/a & n/a \\
			non$\_$occ$\_$discont & 2.7580  &  4.8556 &  n/a & n/a \\
			\hline
		\end{tabular}
		\label{table.e_pep}
	\end{center}
\end{table}

Fig. \ref{fig.eva1} shows the percentage of error pixels with respect to different values of $\gamma_d$. The parameter $\omega_d$ in Eq. (\ref{eq.omega_d}) is a Gaussian weighting function which is dependent on $\gamma_d$. Therefore, the curves in Fig. \ref{fig.eva1} for each region of interest have a local minimum. Similarly, it can be observed that each curve in Fig. \ref{fig.eva2} has a local minimum. This also conforms to the Gaussian weighting function in Eq. (\ref{eq.omega_r}). Finally, it can be seen that for each curve in Fig. \ref{fig.eva3}, the value of $e_{PEP}$ decreases with increasing $\rho$. However, this decrease in $e_{PEP}$ is observed up to the point when $\rho$ is equal to $8$. Beyond the latter value, $e_{PEP}$ increases slightly for each region of interest. This indicates that the performance of the FBS depends on the values of $\omega_d$, $\omega_r$ and $\rho$ and we have to find their optimum values in order to achieve the best performance in terms of accuracy.

\subsection{Speed Evaluation}
\label{sec.speed_eva}

In addition to accuracy, the execution speed of the proposed algorithm is also quantified in order to give an  evaluation of the
overall system's performance. However, due to the fact that image size and disparity range are not constant among different datasets, a general way to depict the
performance in terms of processing speed is given in millions of disparity evaluations per second $Mde/s$ as follows \cite{Tippetts2016}:

\begin{equation}
Mde/s=\frac{u_{max}v_{max}d_{max}}{t}{10^{-6}}
\label{eq.mde_s}
\end{equation}
where $u_{max}$ and $v_{max}$ represent the width and height of the disparity map, $d_{max}$ denotes the maximum search range and $t$ represents the algorithm runtime in seconds. Our implementation achieves a performance of $Mde/s=344.16$  when $\rho$ is set to $6$.

Furthermore, we use the \enquote{Cones} dataset to evaluate the speed performance with respect to different aggregation sizes.
The runtime $t$ corresponding to different values of $\rho$ is shown in Fig. \ref{fig.eva4}. It can be seen that $t$ increases with increasing $\rho$. When $\rho$ is equal to $5$, the execution speed of the proposed implementation is around $50$ fps. Although a better accuracy can be achieved when $\rho$ is set to $7$ (see Fig. \ref{fig.eva3}), the processing speed is doubled (see Fig. \ref{fig.eva4}). Therefore, $\rho=6$ provides a better trade-off between accuracy and speed in the proposed algorithm.

\section{Conclusion and Future Work}
\label{sec.conclusion_fw}

In this paper, we presented the implementation of the FBS on a GTX 1080 GPU. On the algorithm side, the NCC is first simplified as a dot product. The values of $\mu$ and $\sigma$ are pre-calculated and stored in a static program storage for direct indexing. Furthermore, each correlation cost is simultaneously stored in two 3-D cost volumes, thus decreasing the computations by approximately $50\%$ when estimating both the left and right disparity maps. Then, the correlation costs are aggregated adaptively by performing bilateral filtering on the left and right cost volumes. The left and right disparity maps are estimated by simply performing WTA on the filtered cost volumes. Finally, the disparities in half-occluded areas are removed and the subpixel resolution is achieved by conducting parabola interpolations around the initial disparities. By highly exploiting the parallel computing architecture, the proposed FBS performs in real time on the GTX 1080 GPU.

However, performing the bilateral filtering on the whole cost volume is still a time-consuming process. Therefore, we plan to use a group of reliable feature points to suggest the search range for their neighbours. Then, only the correlation costs around the suggested search range are calculated and the bilateral filtering is  performed only on the space around the calculated disparities.

\section{Acknowledgement}
This paper is supported by Shenzhen Science, Technology and Innovation Commission (SZSTI) JCYJ20160428154842603 also supported by the Research Grant Council of Hong Kong SAR Government, China, under Project No. 11210017 and No. 16212815 and No. 21202816, NSFC U1713211 awarded to Prof. Ming Liu; Shenzhen Science, Technology and Innovation Commission (SZSTI) JCYJ20170818153518789 and National Natural Science Foundation of China No. 61603376, awarded to Dr. Lujia Wang.

\bibliographystyle{IEEEbib}

\end{document}